\pdfoutput=1

\documentclass[11pt]{article}
\usepackage{amsmath}
\usepackage[linesnumbered,ruled,vlined]{algorithm2e}
\usepackage[final]{acl}

\usepackage{times}
\usepackage{latexsym}
\usepackage{color}
\usepackage[T1]{fontenc}
\usepackage{longtable}
\usepackage{tabularx}   
\usepackage{ragged2e}   
\usepackage{booktabs}   
\usepackage{enumitem}   
\usepackage{xltabular}  
\usepackage{caption}

\usepackage[utf8]{inputenc}

\usepackage{microtype}

\usepackage{inconsolata}

\usepackage{graphicx}
\usepackage{enumitem}
\usepackage{booktabs}  
\usepackage{multirow}  
\usepackage{array}     
\usepackage{float}
\usepackage{amsmath, amssymb}
\usepackage[utf8]{inputenc}

\title{VERITAS: Leveraging Vision Priors and Expert Fusion to Improve Multimodal Data}

\author{%
Tingqiao Xu\textsuperscript{1}\thanks{Equal contribution.},
Ziru Zeng\textsuperscript{1}\footnotemark[1],
Jiayu Chen\textsuperscript{2}\\
\textsuperscript{1}Shanghai Jiao Tong University \quad
\textsuperscript{2}Fudan University\\
phenomenonkj@sjtu.edu.cn
}

\begin{document}
\maketitle
\begin{abstract}
The quality of supervised fine-tuning (SFT) data is crucial for the performance of large multimodal models (LMMs), yet current data enhancement methods often suffer from factual errors and hallucinations due to inadequate visual perception.  To address this challenge, we propose VERITAS, a pipeline that systematically integrates vision priors and multiple state-of-the-art LMMs with statistical methods to enhance SFT data quality.  VERITAS leverages visual recognition models (RAM++) and OCR systems (PP-OCRv4) to extract structured vision priors, which are combined with images, questions, and answers.  Three LMMs (GPT-4o, Gemini-2.5-Pro, Doubao-1.5-pro) evaluate the original answers, providing critique rationales and scores that are statistically fused into a high-confidence consensus score serving as ground truth.  Using this consensus, we train a lightweight critic model via Group Relative Policy Optimization (GRPO), enhancing reasoning capabilities efficiently.  Each LMM then refines the original answers based on the critiques, generating new candidate answers;  we select the highest-scoring one as the final refined answer.  Experiments across six multimodal benchmarks demonstrate that models fine-tuned with data processed by VERITAS consistently outperform those using raw data, particularly in text-rich and fine-grained reasoning tasks.  Our critic model exhibits enhanced capability comparable to state-of-the-art LMMs while being significantly more efficient.  We release our pipeline, datasets, and model checkpoints to advance research in multimodal data optimization.

\end{abstract}

\begin{figure}[t]
  \includegraphics[width=\columnwidth]{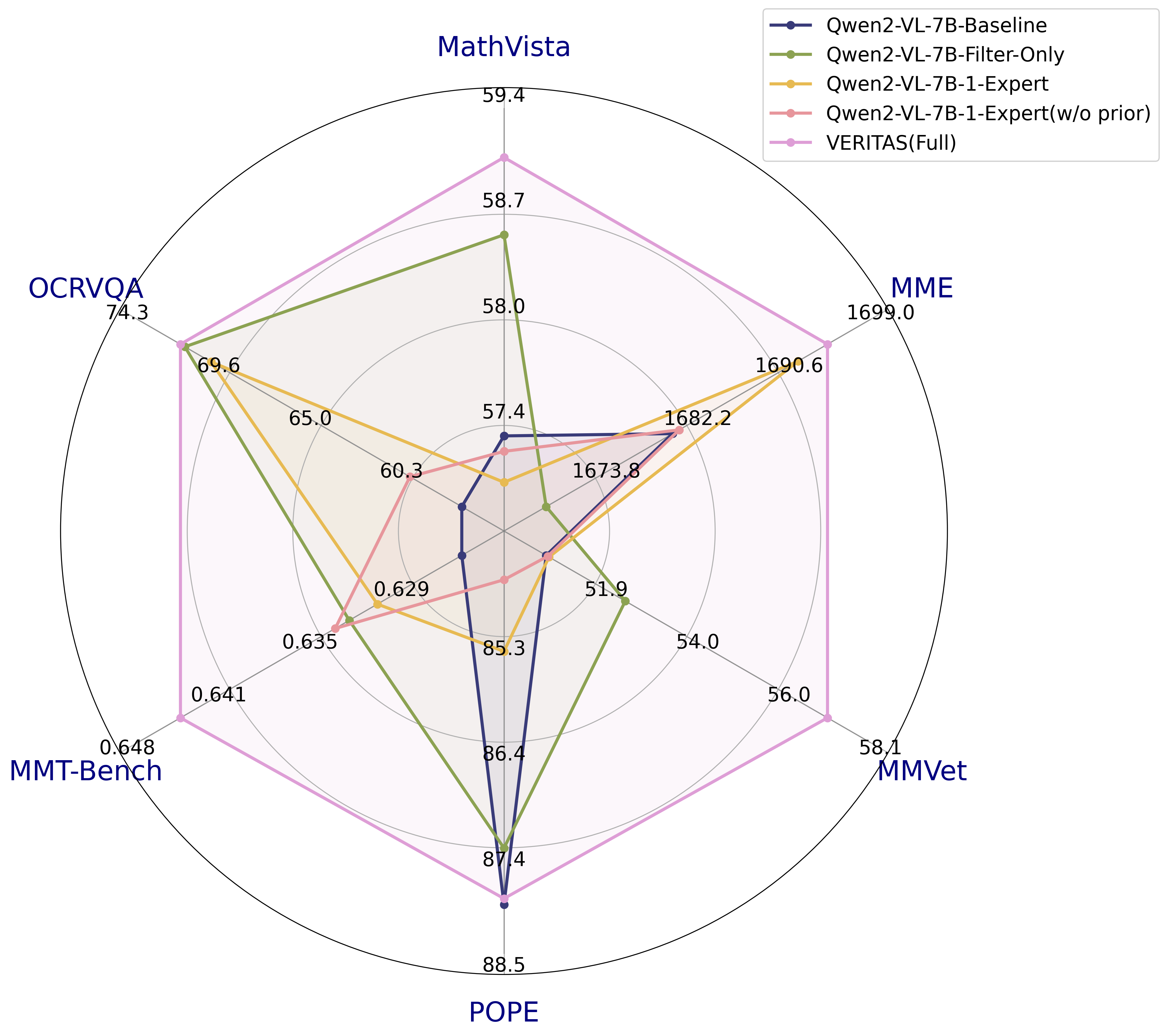}
  \caption{Performance comparison of different models on various benchmarks.}
  \label{fig:radar}
  \vspace{-0.2cm}
\end{figure}

\section{Introduction}
Large multimodal models (LMMs)\citep{chen2024internvl,gpt-4o,gpt-4v,wang2024qwen2,team2024gemini} have recently pushed the frontier of visual–language understanding, yet their ultimate performance is still gated by the quality of the supervised-fine-tuning (SFT) data they learn from\citep{marion2023less, albalak2024survey}. While recent work\citep{luo2024mmevol,liu2024mmdu} enlarges instruction diversity or directly lets a single strong LMM (e.g., GPT-4o\citep{gpt-4o}) synthesize answers \citep{fang2024vila,guo2024mammoth,gu2024infinity}, the generated responses frequently contain factual mistakes, visual hallucinations, or stylistic inconsistencies\citep{bai2024hallucination, liu2024survey, wang2024factuality}. Feeding such noisy data back to SFT not only wastes computation but also hard-limits the attainable accuracy of downstream models.

Two observations motivate this study. First, specialized vision experts such as object detectors and OCR systems remain more reliable than any current LMM on fine-grained perception\citep{zhang2024exploring, fu2024blink}, thus providing trustworthy vision priors. Second, no single LMM can serve as an oracle judge: their preferences are biased, and self-evaluation amplifies their own errors. Therefore, high-quality multimodal data requires (i) external vision priors to ground the scene, (ii) multiple strong but diverse LMM critics to offset individual bias, and (iii) a principled way to fuse these heterogeneous signals at low cost.

We introduce \textbf{VERITAS}, a pipeline for \textbf{V}ision-Priors \textbf{E}valuation and \textbf{R}efinement through \textbf{I}ntegration of \textbf{T}ri-Expert \textbf{A}ssessment with \textbf{S}hrinkage, that systematically upgrades multimodal SFT data through four tightly coupled components:
(1) \textbf{Vision-Prior Extraction} employs RAM++ \citep{huang2023open} and PP-OCRv4 \citep{pp-ocrv4} to convert images into structured tags and texts that are provided to all subsequent modules, effectively anchoring the critique on observable evidence.
(2) \textbf{Tri-Expert Assessment} queries three state-of-the-art LMMs (GPT-4o, Gemini-2.5-Pro\citep{Gemini}, Doubao-1.5-pro\citep{Doubao}) for chain-of-thought critiques and numeric scores. A domain-aware James-Stein–style shrinkage then statistically fuses the three noisy scores into a high-confidence gold score $\hat{S}$, reducing variance without sacrificing unbiasedness.
(3) \textbf{Integration with GRPO} involves training a 7B-parameter multimodal critic using Group Relative Policy Optimization (GRPO)\citep{shao2024grpo}, enabling reasoning-based evaluation of answers in multimodal SFT data for more accurate assessments. By leveraging group-wise advantages, the lightweight critic reproduces GPT-4o–level ranking fidelity while significantly reducing inference costs.
(4) \textbf{Self-Refinement} generates three revised answers conditioned on the vision priors, the expert rationales, and $\hat{S}$. The GRPO Critic selects the best among the original and revised candidates, yielding a final, confidence-graded dataset entry.

Extensive experiments verify the effectiveness of VERITAS. When the same 7B model is SFT-trained on our refined data, it outperforms the counterpart trained on raw data by +7.4 average accuracy over six public benchmarks. The GRPO critic achieves a Kendall $\tau$ of 0.71 with human judgments, only 0.05 behind GPT-4o, yet is two orders of magnitude cheaper to run.
Our contributions are threefold:
\begin{itemize}[leftmargin=*, itemsep=0pt]
\item We propose the first vision-prior+multi-expert scoring framework with domain-aware statistical fusion, theoretically reducing expected risk compared with single-expert or simple averaging baselines.
\item We adapt GRPO to train a lightweight multimodal critic whose ranking consistency rivals GPT-4o at a fraction of the cost, and demonstrate its usefulness for automated answer selection.
\item We release the VERITAS pipeline, the 96K confidence-annotated multimodal dataset, and all model checkpoints to facilitate future research on robust data curation and evaluation.
\end{itemize}

\begin{figure*}[t]
  \includegraphics[width=1\linewidth]{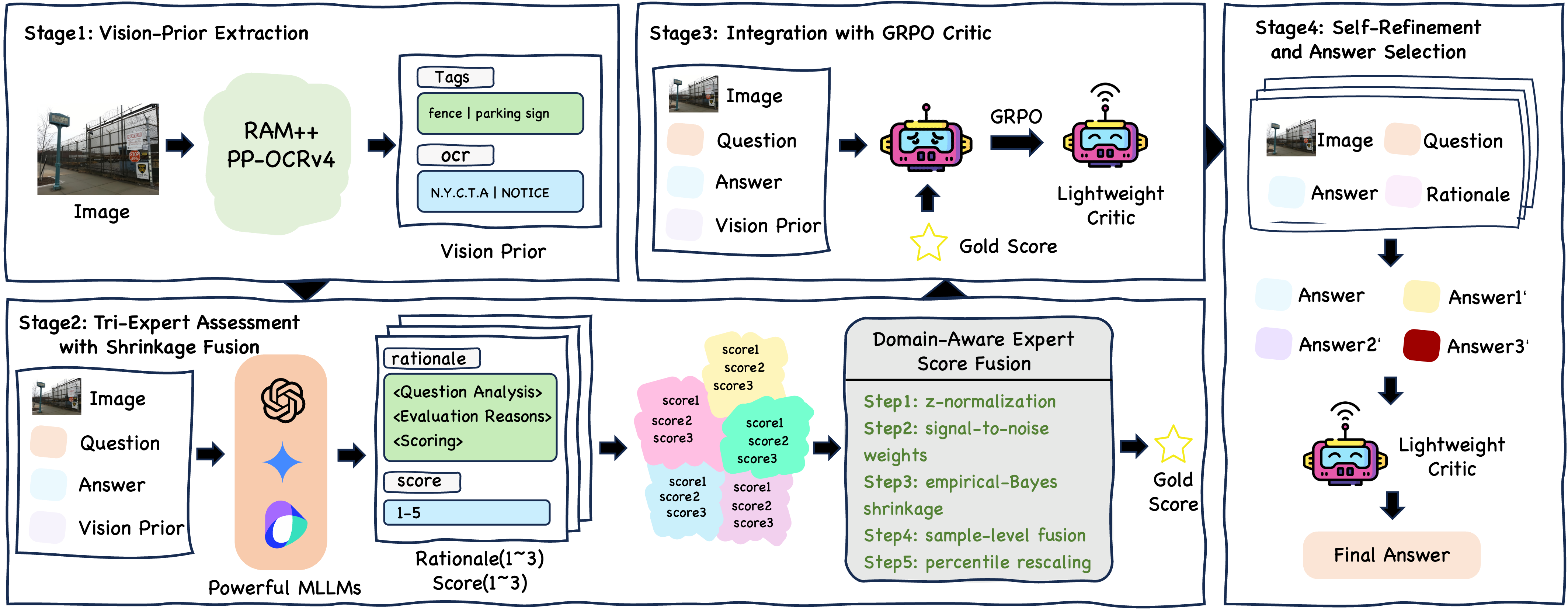}
  \caption {Overview of the proposed pipeline VERITAS containing four stages: (1) Vision-Prior Extraction, (2) Stage2: Tri-Expert Assessment with Shrinkage Fusion, (3) Integration with GRPO Critic, (4) Self-Refinement and Answer Selection.}
  \label{fig:pipeline}
  \vspace{-0.2cm}
\end{figure*}

\section{Related Work}

\noindent \textbf{Advancements in Multimodal Evaluation and Critique}\quad
Evaluating multimodal models poses significant challenges due to the intricate interplay between visual and textual modalities. Traditional text-based metrics fail to capture the complexity of visual information~\citep{zheng2023judging}. The emergence of the ``LLM-as-a-judge'' paradigm, where large language models (LLMs) serve as evaluators, has brought transformative changes. Recent innovations include methods like EvalPlanner~\citep{saha2025learningtoPlan}, which decomposes the evaluation process into planning and reasoning stages, employing a self-training loop with supervised fine-tuning and Direct Preference Optimization (DPO) to enhance the evaluator's capabilities. Another approach, Self-Generated Critiques~\citep{yu2024Self-generated-critiques}, leverages a model's own critiques to bolster reward modeling, providing fine-grained feedback that reduces reward hacking risks and enhances interpretability. Similarly, Generative Verifiers~\citep{zhang2024generative-verifiers} reframe reward modeling as next-token prediction, utilizing the generative strengths of language models to assess input quality without direct scalar scoring.

In the multimodal context, models like R1-Reward~\citep{zhang2025R1-Reward} employ reinforcement learning to train multimodal reward models, achieving significant improvements on benchmarks like VL RewardBench and MM Reward Bench. However, these methods often rely on single-model evaluations, which can introduce biases and accumulate errors, particularly due to the limitations of LMMs in fine-grained visual perception~\citep{shumailov2023curse, zhang2024exploring}. There is a need for approaches that integrate vision priors from specialized models with assessments from multiple expert LMM critics, employing statistical fusion methods to enhance the reliability of evaluations while reducing computational costs.

\noindent \textbf{Multimodal Data Refinement and Self-Improvement Mechanisms}\quad
Enhancing the quality of multimodal training data is essential for advancing model performance. Prior efforts like CritiqueMM~\citep{ke2024critiquellm} and VILA~\citep{fang2024vila} utilize iterative self-critique processes, where LMMs refine data using their own feedback. However, the inherent limitations of LMMs in visual understanding can lead to error propagation and persistent issues like hallucinations~\citep{zhang2024exploring,shumailov2023curse}. We address these problems by integrating vision priors, multi-expert feedback, and a refinement mechanism to enhance multimodal data quality without error accumulation.

In summary, a significant research gap exists in systematically integrating domain-specific vision experts with LMMs to enhance multimodal data critique and refinement. Moreover, existing critic models often rely on single-model evaluations without harnessing the benefits of multi-expert feedback and statistical confidence fusion. Addressing these issues could overcome current model limitations and improve the quality of supervised fine-tuning data.

\section{Method}
In this section, we propose \textbf{VERITAS}, a lean vision-prior + multi-expert pipeline for automatic data critique and refinement.  The workflow has four concise steps:
(1) \textbf{Vision-Prior Extraction}.  RAM++ and PP-OCRv4 turn each image into object tags and OCR text, providing grounded evidence.  
(2) \textbf{Tri-Expert Assessment with Shrinkage Fusion}.  Three strong MLLMs (GPT-4o, Gemini-2.5-Pro, Doubao-1.5-pro) critique each answer with the priors; a domain-aware James–Stein shrinkage merges their scores into a single high-confidence label $\hat S$.  
(3) \textbf{ Integration with GRPO Critic}.  We distil the costly ensemble into a 7B critic via Group Relative Policy Optimisation, retaining GPT-4o-level ranking at a fraction of the cost.  
(4) \textbf{Self-Refinement and Answer Selection}.  The experts rewrite the answer; the GRPO critic selects the best candidate, yielding a confidence-graded, denoised dataset.

This compact design grounds vision, mitigates single-model bias, and scales high-quality multimodal SFT data.

\subsection{Data Collection}

To validate model generalization capabilities on heterogeneous data, we systematically collected 7 benchmark datasets covering 6 core visual-language tasks: Fine-grained image caption leverages Image Textualization \citep{pi2024imageTextualization} and TextCaps \citep{sidorov2020textcaps}; LLaVAR \citep{zhang2023llavar} enhances text-rich image understanding; Domain-specific reasoning employs QA pairs from AI2D(GPT4V) \citep{li2024llava-onevision}; Complex reasoning synthesizes instances from ShareGPT4V \citep{chen2024sharegpt4v}; Multi-turn dialogue and reasoning modeling utilizes LRV-Normal \citep{li2024llava-onevision} for long-term context tracking; Hallucination mitigation integrates samples in LRV-Instruction \citep{liu2023aligning}. 

The final dataset comprises 96K samples, comprehensive statistics are provided in Table \ref{tab:data_composition}.

\begin{table}
  \centering
  \setlength{\tabcolsep}{10pt}
  \begin{tabular}{lc}
    \hline
    \textbf{Dataset} & \textbf{Samples} \\
    \hline
     Image Textualization & 14,825 \\
     TextCaps             & 11,014 \\ 
     LLaVAR               & 19,774 \\ 
     AI2D (GPT4V)         & 4,864  \\
     ShareGPT4V           & 15,001 \\ 
     LRV-Normal           & 10,477 \\
     LRV-Instruction      & 20,000 \\ 
     \hline
     \textbf{Total} & \textbf{95,955} \\
     \hline
  \end{tabular}
  \caption{Data Collection Statistics. In total, 95,955 samples were gathered.}
  \label{tab:data_composition}
  \vspace{-0.2cm}
\end{table}

\subsection{VERITAS: A Four-Stage Pipeline for Data Critique and Refinement}
To move beyond a single-critic setting, VERITAS decomposes data curation into four successive stages, each adding an orthogonal source of reliability. Figure \ref{fig:pipeline} gives an overview.

\noindent\textbf{Stage 1: Vision-Prior Extraction}\quad  
We first ground every sample with explicit perceptual evidence. Two off-the-shelf specialist models are invoked on the input image $I$:
\begin{align}
\text{Tags} &= \mathrm{RAM^{++}}(I), \\
\text{OCR}  &= \mathrm{PP\text{-}OCRv4}(I).
\end{align}
The resulting object labels and text strings are serialised as a string prior  
$V=\{\text{Tags},\text{OCR}\}$ and appended to all subsequent prompts using a natural-language wrapper.

\noindent\textbf{Stage 2: Tri-Expert Assessment with Shrinkage Fusion}\quad

\textit{Multi-Expert Critique.} Three state-of-the-art MLLMs (GPT-4o, Gemini-2.5-Pro, and Doubao-1.5-pro) independently assess answer quality, with each model $m$ producing:
\begin{equation}
(s_m, r_m) = \mathcal{M}^{(m)}_{\text{critic}}(I, q, a_0, V \mid \mathcal{E}_c)
\end{equation}
where $s_m \in [0,5]$ is a scalar score and $r_m$ is a structured rationale following our evaluation rubric $\mathcal{E}_c$. Each expert receives the same vision priors extracted in Stage 1, ensuring grounded assessments.(The complete prompt template is shown in Appendix \ref{sec:Critique-Prompt})

\textit{Domain-Aware Score Fusion.} Three MLLMs produce raw scores  $s_{m}(n)\in[0,5]$.  We transform them into a single confidence $\hat{S}(n)$ through the following steps.
(1) z-normalise each critic inside its domain;  
(2) compute a domain-wise signal-to-noise ratio (SNR) and use it as a weight; 
(3) shrink that weight toward a corpus prior via an empirical-Bayes factor \(\alpha_{d}=N_{d}/(N_{d}+\lambda)\);  
(4) form the weighted average of the normalised scores;  
(5) map the result back to the 0–5 rubric by a robust 5 \%–95 \% percentile stretch. The specific process is presented in Algorithm \ref{alg:fusion}

\noindent\textbf{Stage 3: Integration with GRPO Critic}\quad  
We distil fused judgement Ŝ into a 7B parameter critic using \emph{Group Relative Policy Optimisation} (GRPO).  GRPO eliminates an extra value–function by comparing the candidate answer with a \textit{group baseline} drawn from the same old policy, thereby producing a low-variance, self-normalising advantage.
\vspace{2pt}

\textit{Training objective.}
For every image–question pair $q$, we sample a group of $G$ drafts $\{o_i\}_{i=1}^{G}$ from the policy $\pi_{\theta_{\text{old}}}$.  The new policy $\pi_{\theta}$ is updated by maximising
\begin{equation}
\begin{aligned}
\mathcal{J}_{\text{GRPO}}(\theta) ={} & \mathbb{E}_{q,\{o_i\}}\, \frac{1}{G} \sum_{i=1}^{G} \frac{1}{|o_i|} \sum_{t=1}^{|o_i|} \Biggl[\\ \min\Bigl( \dfrac{\pi_{\theta}^{i,t}}{\pi_{\theta_{\text{old}}}^{i,t}}\hat{A}_{i,t}, 
&  \text{clip}\bigl(\dfrac{\pi_{\theta}^{i,t}}{\pi_{\theta_{\text{old}}}^{i,t}},1-\epsilon,1+\epsilon\bigr) \hat{A}_{i,t} \Bigr) \\
& - \beta\, \mathrm{D}_{\text{KL}} \bigl( \pi_{\theta} \| \pi_{\text{ref}} \bigr) \Biggr],
\end{aligned}
\end{equation}
where $\hat A_{i,t}$ is the group–relative advantage
\vspace{2pt}

\textit{Reward design.}  Each rollout receives two additive rewards  
$R_i=R_{\text{acc}}+R_{\text{fmt}}$:

• Accuracy reward ($R_{\text{acc}}$) compares the scalar score extracted from $o_i$ with the Stage-2 “gold” $\hat S$  
\begin{equation}
R_{\text{acc}}
=\max\!\Bigl(0,1-\frac{|\text{int}(o_i)-\hat S|}{5}\Bigr).
\end{equation}

• Format reward ($R_{\text{fmt}}$) encourages the critic style we need downstream:  
\begin{equation}
\begin{aligned}
N = 
& \#\bigl( \texttt{<Question Analysis>} \in o_i \bigr) +\\
& \#\bigl( \texttt{<Evaluation Reasons>} \in o_i \bigr) + \\
& \#\bigl( \texttt{<Scoring>} \in o_i \bigr)
\end{aligned}
\end{equation}

\begin{equation}
R_{\text{fmt}} = 0.5 \times \left( \dfrac{N}{3} \right)
\end{equation}

The indicator function $\#\bigl(s_k \in o_i\bigl)$ equals $1$ if $s_k$ appears in $o_i$, and $0$ otherwise

\begin{algorithm}[htbp]
\caption{Domain-Aware Expert Score Fusion}
\label{alg:fusion}
\DontPrintSemicolon
\small
\SetKwData{Left}{left}\SetKwData{This}{this}\SetKwData{Up}{up} \SetKwFunction{Union}{Union}\SetKwFunction{FindCompress}{FindCompress} \SetKwInOut{Input}{input}\SetKwInOut{Output}{output}
	
\Input{Scores \(s_{m}(n)\), \(m = 1..3\); domain labels \(d(n)\); constants \(\epsilon = 10^{-3}\), \(\lambda = 100\)}
\Output{Fused confidence \(\hat S(n) \in [0, 5]\)}
\BlankLine
\textbf{Step 0: domain statistics}
\For{\(m \gets 1 \textbf{ to } 3\)}{
    \For{\(d \in \mathcal{D}\)}{
        \(\mu_{m,d} \gets \text{mean}(s_m(n) \mid d(n) = d)\)\;
        \(\sigma_{m,d} \gets \text{std}(s_m(n) \mid d(n) = d)\)\;
    }
}
\BlankLine
\textbf{Step 1: z-normalisation}
\ForEach{\text{samples \( n \)}}{
    \For{\(m \gets 1 \textbf{ to } 3\)}{
        \(z_m(n) \gets \frac{s_m(n) - \mu_{m,d(n)}}{\sigma_{m,d(n)} + \epsilon}\)\;
    }
}
\BlankLine
\textbf{Step 2: signal-to-noise weights}
\For{\(m \gets 1 \textbf{ to } 3\)}{
    \For{\(d \in \mathcal{D}\)}{
        \textit{Consensus}\(_{d} \gets \text{mean}_{k}(s_{k} \mid d)\)\;
        \(r_m(n) \gets s_m(n) - \textit{Consensus}_{d}\) \;
        \textit{Sig}\(_{m,d} \gets \text{std}(s_m \mid d)\)\;
        \textit{Noise}\(_{m,d} \gets \text{std}(r_m \mid d)\)\;
        \textit{Raw weight}\(_{m,d} \gets \frac{\textit{Sig}_{m,d}}{\textit{Noise}_{m,d} + \epsilon}\)\;
    }
}
\BlankLine
\textbf{Step 3: empirical-Bayes shrinkage}
\For{\(m \gets 1 \textbf{ to } 3\)}{
    \(\overline{w}_m \gets \text{mean}_{d}(\textit{Raw weight}_{m,d})\)\;
}
\For{\(d \in \mathcal{D}\)}{
    \(\alpha_d \gets \frac{N_d}{N_d + \lambda}\)\;
    \For{\(m \gets 1 \textbf{ to } 3\)}{
        \(\hat{w}_{m,d} \gets \alpha_d \cdot \textit{Raw weight}_{m,d} + (1 -
       \alpha_d) \cdot \overline{w}_m\)\;
    }
    Normalize: \(\hat{w}_{m,d} \gets \frac{\hat{w}_{m,d}}{\sum_{k=1}^{3}\hat{w}_{k,d}}\)\;
}
\BlankLine
\textbf{Step 4: sample-level fusion}
\ForEach{\text{samples \( n \)}}{
    \(\hat{z}(n) \gets \sum_{m=1}^{3}\hat{w}_{m,d(n)} \cdot z_m(n)\)\;
}
\BlankLine
\textbf{Step 5: percentile rescaling}
\((q_{\text{low}}, q_{\text{high}}) \gets\) 5\% / 95\% quantiles of \(\{\hat{z}(n)\}\)\;
\ForEach{\text{samples \( n \)}}{
    \(\hat{S}(n) \gets 5 \cdot \text{clip}\left(\frac{\hat{z}(n) - q_{\text{low}}}{q_{\text{high}} - q_{\text{low}}}, 0, 1\right)\)\;
}
\Return \(\hat{S}(n)\)\;
\end{algorithm}
(A full derivation and risk analysis are deferred to Appendix \ref{sec:Derivation-Multi-Expert}.)

\noindent\textbf{Stage 4: Self-Refinement and Answer Selection}\quad 
The rationales $\{r_m\}$ and score $\hat S$ serve as fine-grained feedback for rewriting(The complete prompt template is shown in Appendix \ref{sec:Refinement-Prompt}):
\begin{equation}
a_m'=\mathcal M_{\text{rewrite}}^{(m)}(I,q,a_0,r_m,\hat S\mid\mathcal E_r).
\end{equation}
We form a candidate pool
$\mathcal C=\{a_0,a_1',a_2',a_3'\}$
and ask the GRPO critic to rescore each candidate:
\begin{equation}
\tilde s=\mathcal M_{\text{GRPO}}(I,q,a,V),\quad a\in\mathcal C.
\end{equation}
The highest-ranked answer
$\hat a=\arg\max_{a\in\mathcal C}\tilde s$
is retained together with its confidence $\hat S$ and a merged rationale $\bar r$.  The refined dataset
$\mathcal D_{\text{refine}}=\{(I,q,\hat a,\hat S,\bar r)\}$
achieves substantial quality gains without shrinking in size, overcoming the severe recall loss of rigid threshold filtering.

This four-stage design grounds each judgement in observable evidence, balances multiple expert opinions through principled statistics, amortises cost via a compact critic, and finally delivers high-confidence, hallucination-free multimodal supervision. A complete example of the entire process is provided in Appendix \ref{sec:appendix-case-study}

\section{Experimental Setup}\label{sec:setup}
This section describes all data resources, model configurations, and evaluation protocols used in our study. 
\subsection{Training Corpora}\label{sec:train-data}
\paragraph{VERITAS Instruction Corpus.}  
Table~\ref{tab:data_composition} lists seven public multimodal instruction sources that form our base corpus.  For each source we keep the original \textsc{Raw} answer and the \textsc{Refine} answer produced by the VERITAS pipeline, resulting in two parallel sets($\mathcal D_{\text{raw}}$ and $\mathcal D_{\text{refine}}$) of identical size (95,955 samples each).

\subsection{Critic-Evaluation Sets}\label{sec:critic-data}
\paragraph{In-domain 1K.}  
To assess in-domain ranking fidelity, we randomly sample 1,000 image–question–answer triplets from the same seven sources while ensuring no overlap with the training split. Three human annotators independently assign an integer quality score from 0 (worst) to 5 (best); majority vote is taken as the reference label.
\paragraph{Out-of-domain CLEVR-500.}  
We further probe generalisation with 500 images from the CLEVR\citep{johnson2017clevr} test split.  All original answers are correct (\emph{good}).  We automatically inject (i) minor attribute swaps (\emph{medium}) and (ii) severe object-count or colour mistakes (\emph{bad}) so that the final distribution is 160 / 170 / 170. Three human annotators assign ground-truth scores in the same manner as above. Full injection rules are provided in Appendix \ref{sec:Error-Injection}.

\subsection{Down-Stream Benchmarks}\label{sec:benchmarks}
We adopt six widely-used public benchmarks that cover perception, reasoning, and hallucination:
\begin{itemize}[leftmargin=*,nosep]
\item \textbf{MME} \citep{mme}: 14 binary diagnostics for basic perception.  
\item \textbf{OCR-VQA} \citep{ocrvqa}: text-in-image understanding.  
\item \textbf{MM-Vet} \citep{mmvet}: open-ended evaluation across 16 capabilities.  
\item \textbf{MathVista} \citep{mathvista}: visual-symbolic mathematical reasoning.  
\item \textbf{MMT-bench} \citep{mmtbench}: 32 meta-tasks including autonomous driving. 
\item \textbf{POPE} \citep{li2023POPE}: hallucination detection.
\end{itemize}
All evaluations are conducted using the Vlmevalkit \citep{duan2024vlmevalkit} toolkit.

\subsection{Model and Training Details}\label{sec:training}
\paragraph{Model Architecture.}  
Every model in this study is based on Qwen2-VL-7B.  We load the \textsc{Qwen2-VL-Instruct} checkpoints as initialisation and keep the vision encoder frozen throughout.
\paragraph{Instruction Fine-tuning.}  
Models trained on the Raw and Refine splits of the VERITAS Instruction Corpus are each fine-tuned for one epoch with learning rate 5e-6, batch size 64, and cosine decay. The identical hyper-parameter configuration is applied to every ablation variant to ensure a fair comparison.
\paragraph{Lightweight GRPO critic.}  
For the lightweight critic, we finetune another Qwen2-VL-7B using GRPO. Before commencing GRPO training, we first performed a "cold start" training using 6,000 data samples. Each update samples $G{=}128$ candidate rationales, and the total training lasts one epoch over the 95,955 fused score items.  We set $\beta{=}0.01$ for the KL term and clip ratio $\epsilon{=}0.2$.

\subsection{Baselines and Ablations}\label{sec:ablationsetup}
To isolate the contribution of every component, we instantiate seven SFT variants:
\begin{enumerate}[leftmargin=*,nosep,label=(\alph*)]
\item \textbf{Raw}: trained on $\mathcal D_{\text{raw}}$.  
\item \textbf{VERITAS}: full pipeline, trained on $\mathcal D_{\text{refine}}$.  
\item \textbf{1-Expert\,+VP}: one LMM critic + vision priors, no fusion.  
\item \textbf{1-Expert}: one LMM critic, no vision priors.  
\item \textbf{Filter-Only}: remove samples with $\hat S\!<\!\tau$ (keep $\approx$50 K), no rewriting.
\item \textbf{VERITAS(w/o fusion)}: replace shrinkage fusion with the average value.
\item \textbf{VERITAS(open-source)}: replace the closed-source ex-
pert trio with open-source models.
\end{enumerate}
For the critic task we compare two systems:
\begin{itemize}[leftmargin=*,nosep]
\item \textbf{Lightweight GRPO critic}: our distilled critic.  
\item \textbf{Lightweight SFT critic}: supervised fine-tuning on GPT-4o scores and rationales only.  
\end{itemize}

\section{Results and Discussion}\label{sec:results}

We first present the impact of the proposed pipeline on downstream task accuracy (§\ref{sec:overall}), then isolate the effect of each design choice through ablations (§\ref{sec:ablation}).
The quality and efficiency of the lightweight \textit{GRPO critic} are analysed in §\ref{sec:critic}, while §\ref{sec:fusion} examines why the domain–aware fusion is superior to naive scoring.

\begin{table*}
  \centering
  \setlength{\tabcolsep}{6pt}
  \begin{tabular}{@{}l*{5}{c}c@{}}
    \hline
    \textbf{Model} & \textbf{MME} & \textbf{OCR-VQA} & \textbf{MM-Vet} & \textbf{MathVista} & \textbf{MMT-bench} & \textbf{POPE}\\
    \hline
    Baseline                & 1680.9 & 57.780 & 50.780 & 57.3 & 0.625 & \textbf{87.97}   \\
    Filter-Only             & 1669.3 & 71.908 & 52.569 & 58.6 & 0.633 & 87.41  \\ 
    1-Expert                & 1692.4 & 70.573 & 50.844 & 57.0 & 0.631 & 85.46  \\ 
    1-Expert(w/o prior)     & 1681.5 & 60.424 & 50.814 & 57.2 & 0.629 & 84.75  \\ 
    VERITAS(w/o fusion)  & 1694.2 & 70.922 & 55.400 & 57.6 & 0.633 & 86.31  \\
    VERITAS(open-source)  & 1686.1 & 64.323 & 50.240 & 56.6 & 0.625 & 85.28  \\
    VERITAS\textbf{(Full)}  & \textbf{1695.1} & \textbf{72.133} & \textbf{57.142} & \textbf{59.1} & \textbf{0.645} & 87.91  \\ 
    \hline
  \end{tabular}
  \caption{Overall performance comparison of models trained under different configurations on various benchmarks. The best performance for each benchmark and model size is highlighted in bold.}
  \label{tab:overall_performance}
  \vspace{-0.2cm}
\end{table*}

\subsection{Overall Down-stream Performance}\label{sec:overall}

Table \ref{tab:overall_performance} summarises the accuracy of seven SFT variants on six public benchmarks. 

\paragraph{Large gains on perception‐centric tasks.}  
VERITAS delivers the strongest improvements on the two perception-heavy suites, \textbf{OCR-VQA} (+14.35) and \textbf{MME} (+14.2).  
Because these tasks require precise localisation of characters, symbols and fine details, the explicitly injected \emph{vision priors} (object tags + OCR strings) provide grounded evidence that the model can directly reference during supervision.  The effect is already visible in the “1-Expert” ablation, and becomes maximal when multi-expert fusion and rewrite are enabled.

\paragraph{Cross-modal understanding.}  
On \textbf{MM-Vet} the full pipeline surpasses the baseline by +6.36 points, markedly higher than any single ablation.  The test set combines 16 question styles (e.g., attribute comparison, positional reasoning).  We conjecture that (i) denoising removes contradictory rationales, and (ii) shrinkage fusion supplies a better-calibrated score gradient for learning nuanced cross-modal associations.

\paragraph{Hallucination behaviour.}  
On \textbf{POPE} the hallucination rate of VERITAS (87.91 is better) is statistically on par with the baseline (87.97).  The negligible -0.06 difference indicates that rewriting does \emph{not} introduce new hallucinations, while the vision priors prevent the critic from falsely rewarding unsupported statements.  Compared with the "1-Expert(w/o prior)" ablation (–3.2) and the "1-Expert" ablation (–2.5), vision priors and multi-expert verification clearly reduce hallucination risk.

\subsection{Ablation Studies}\label{sec:ablation}
\begin{figure}[t]
  \includegraphics[width=1\linewidth]{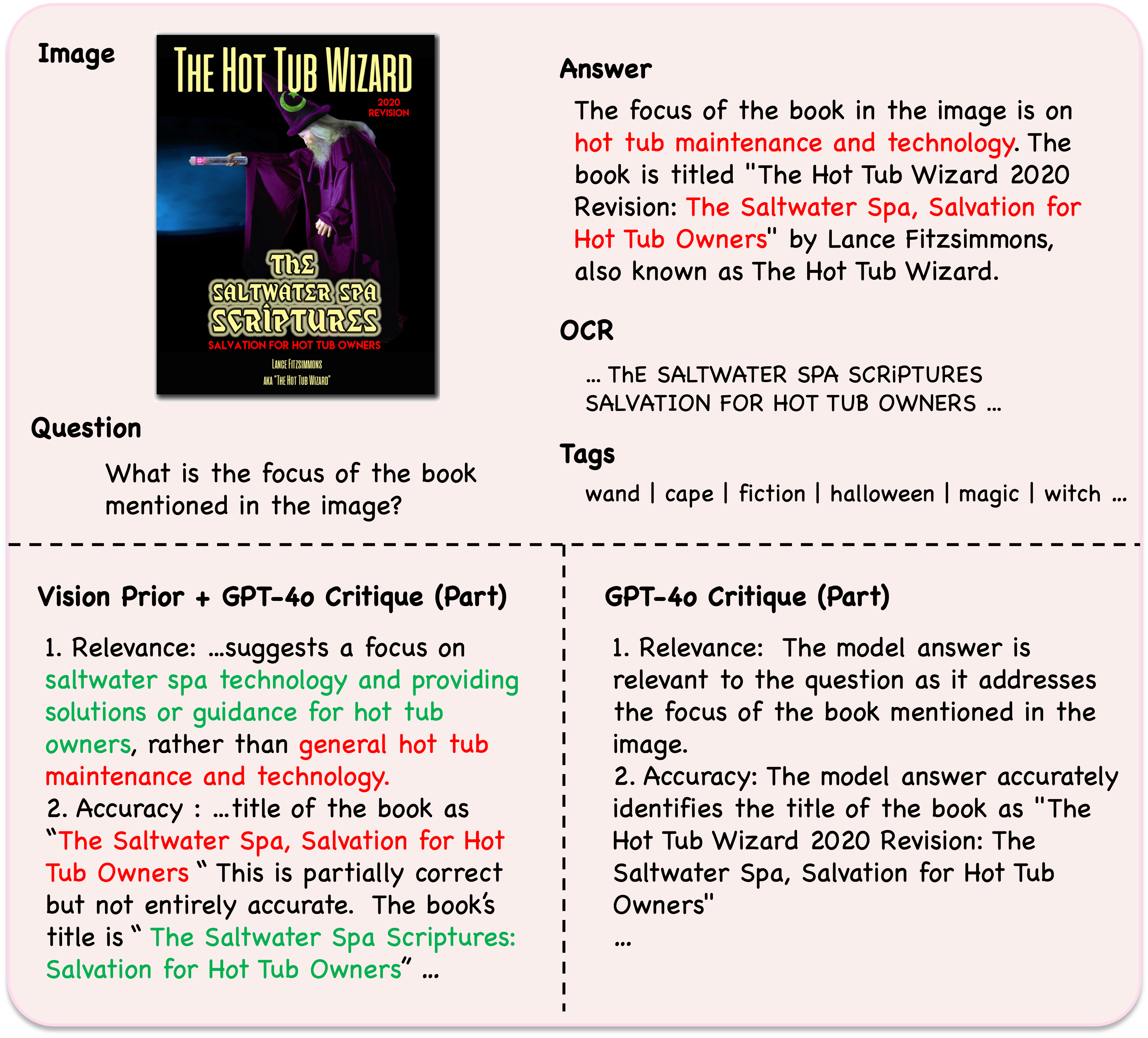}
  \caption {A Comparative Analysis of Critique Before and After the Integration of Vision Prior in GPT-4o. The sections marked in red denote inaccuracies in the answer and also reflect the critique model's identification of these errors, while the sections highlighted in green represent accurate critique.}
  \label{fig:vision-prior-case}
  \vspace{-0.2cm}
\end{figure}

\paragraph{Impact of Vision Priors}\label{sec:ablation_vp}
Comparing \textsc{1-Expert} with \textsc{1-Expert(w/o prior)} isolates the effect of importing object tags and OCR strings from external detectors.
The prior raises \emph{OCR-VQA} by +10.1 points and \emph{MME} by +10.9, while leaving general reasoning tasks largely unchanged.
Such gains confirm that the critic benefits from explicit low-level evidence when judging fine-grained answers; without it, the single-expert critic often fails to spot subtle transcription or attribute errors that state-of-the-art LMMs occasionally make.
The modest improvement on \emph{POPE} (+0.7) suggests that the prior also mitigates hallucinations stemming from non-existent text or objects.
An illustrative example demonstrating how the vision prior aids the answer critique is presented in Figure \ref{fig:vision-prior-case}.

\paragraph{Multi-Expert Scoring and Diversity Driven Rewriting}\label{sec:ablation_multi}

To quantify the contribution of multi-expert scoring, we contrast \textsc{VERITAS (Full)} with the single-expert variant \textsc{1-Expert}.  
Adding two further experts increases the number of (score,rationale) pairs from one to three; consequently, the rewriter produces three alternative answers, and the GRPO critic selects the best among four candidates in total.  

As reported in Table~\ref{tab:overall_performance}, this additional diversity yields consistent improvements on four of the six benchmarks:  
MM-Vet (+6.30), MME (+2.70), OCR-VQA (+1.37), and MathVista (+1.8). Because MM-Vet comprises 16 heterogeneous question types (attribute comparison, positional reasoning, etc.), the large margin indicates that critics with different inductive biases expose complementary error patterns that a single expert fails to uncover.  The GRPO critic can therefore select higher-quality rewrites from a richer hypothesis space, translating into measurable downstream gains.

\paragraph{GRPO Critic: Selecting among Rewrites vs.\ Discarding Data}\label{sec:ablation_grpo}
A hard-filter baseline (\textsc{Filter-Only}) removes low-scored items, which improves OCR-VQA performance but eliminates 46 \% of the training set, thereby limiting the diversity and richness of the training data and lowering MME by 11.6 points.
Our alternative keeps coverage: for every filtered sample we generate three expert-guided rewrites, then let the GRPO critic rescore the original plus the rewrites and keep the top candidate. 

Empirically, GRPO selection outperforms hard filtering on all tasks: relative to \textsc{Filter-Only}, the full VERITAS pipeline gains +25.8 on MME and +4.6 on MM-Vet while remaining parity on hallucination (POPE).  
Thus, choosing the best among multiple rewrites yields a markedly better quality–quantity trade-off than simply discarding noisy samples.

\paragraph{Impact of Shrinkage Fusion}\label{sec:ablation_shrink}
Compared to mean-and-round averaging in VERITAS(w/o fusion), the domain-aware shrinkage fusion in VERITAS(Full) yields consistent improvements on key suites, notably around +1.2 on OCR-VQA and +1.7 on MM-Vet, while also lowering hallucinations as reflected on POPE. We attribute these gains to better cross-domain calibration via per-domain z-normalization and SNR-based weighting, variance reduction from James–Stein–style shrinkage (especially in low-data domains), and the preservation of continuous supervision targets that avoid quantization noise. Together, these factors provide a smoother and more faithful training signal for the GRPO critic and lead to more reliable candidate selection during refinement.

\paragraph{Scalability and Portability with Open-Source Expert Trios}\label{sec:ablation_experts}
Replacing the closed-source expert trio with open-source models—Qwen-2.5-VL-72B, InternVL-3-78B, and Ovis2-34B—while keeping the same VERITAS pipeline (3-Expert, open-source) still delivers clear gains over the baseline, notably on text-rich perception and fine-grained recognition suites such as OCR-VQA and MME, demonstrating the method’s portability. However, this open-source variant trails VERITAS(Full) on reasoning-heavy and hallucination-sensitive benchmarks, indicating that the ultimate performance ceiling remains constrained by the strength of the open-source expert models(shown in Table \ref{tab:critic_correlation}), stronger critics enable the pipeline to realize its full potential.

\subsection{Critic Quality and Efficiency}\label{sec:critic}
Table~\ref{tab:critic_correlation} investigate how well each critic reproduces human judgements and how robust that behaviour remains under a domain shift. 

\begin{table*}
  \centering
  \setlength{\tabcolsep}{6pt}
  \begin{tabular}{lcccc}
    \toprule
    \multirow{2}{*}{\textbf{Model}} & \multicolumn{2}{c}{\textbf{In-Domain}} & \multicolumn{2}{c}{\textbf{Out-Domain}}\\
    \cmidrule(lr){2-3}\cmidrule(l){4-5}
     & Pearson-r & Kendall’s Tau & Pearson-r & Kendall’s Tau\\
    \midrule
    Qwen2-VL-7B-Instruct            & 0.122 & 0.078 & 0.165 & 0.080\\
    InternVL-3-78B                  & 0.421	& 0.410	& 0.427	& 0.422\\
    GPT-4o                          & 0.816 & 0.761 & 0.822 & 0.773\\
    Lightweight SFT critic (ours)   & 0.689 & 0.676 & 0.312 & 0.278\\
    Lightweight GRPO critic (ours)  & \textbf{0.724} & \textbf{0.711} & \textbf{0.628} & \textbf{0.601}\\
    \bottomrule
  \end{tabular}
  \caption{Correlation between critic scores and human scores (higher is better).  GRPO critic outperforms the SFT critic and approaches GPT-4o while being two orders of magnitude cheaper.}
  \label{tab:critic_correlation}
\end{table*}

\paragraph{In-domain Evaluation.}  
On the 1K dev set drawn from the same seven public data sources, the naive Qwen2 baseline is essentially uncorrelated with human raters ($r{=}0.12$).  Both distillation methods greatly narrow the gap to GPT-4o: the SFT critic reaches $r{=}0.689$, while GRPO pushes the figure to $0.724$ and Kendall’s~$\tau$ to $0.711$, i.e.\ \textbf{89 \%} of GPT-4o’s fidelity—already competitive for practical data curation.

\paragraph{Out-of-Domain Robustness.} 
The performance gap becomes more pronounced in the out-of-domain CLEVR dataset. The \textbf{Lightweight GRPO Critic} maintains a reasonable correlation (Pearson $r=0.628$, Kendall's $\tau=0.601$), demonstrating robust generalization to unseen data distributions. In contrast, the \textbf{Lightweight SFT Critic} experiences a significant drop in correlation (Pearson $r=0.312$, Kendall's $\tau=0.278$), indicating overfitting to the in-domain data and poor transferability.

We attribute this to the \emph{relative} objective of GRPO, which forces the policy to model fine-grained ranking differences rather than absolute score regression, making it less sensitive to distributional shifts in raw score ranges.

\subsection{Effectiveness of Domain-Aware Fusion}\label{sec:fusion}

\paragraph{Distributional calibration.}
Figure~\ref{fig:violin_scores} illustrates the score distributions assigned by the individual critics—GPT-4o, Doubao-1.5-Pro, and Gemini-2.5-Pro—compared to the distribution after applying domain-aware fusion. The individual critics exhibit varying scoring tendencies and our fusion method recalibrates these discrepancies, resulting in a more balanced and unimodal distribution that better reflects the true quality of the data. This demonstrates that domain-aware fusion effectively combines the strengths of individual critics while mitigating their biases, leading to more reliable and consistent scoring across the dataset.

\paragraph{Adaptation of critic weights across domains.}
Figure~\ref{fig:critic_weights} shows the changes in critic weights before (raw weights) and after fusion (fused weights) across the seven data sources. The raw weights indicate the initial influence of each critic, while the fused weights demonstrate how the domain-aware fusion adjusts these weights based on the reliability of each critic in different domains. Notably, in data source 5, the weight for Gemini-2.5-Pro increases significantly while Doubao-1.5-Pro's weight decreases, reflecting Gemini's stronger performance and reliability in that specific domain. This adaptive weighting enhances the overall scoring accuracy by leveraging each critic's strengths where they are most effective, leading to improved data quality for downstream tasks.

\section{Conclusion}

In this paper, we introduced \textbf{VERITAS}, a comprehensive pipeline designed to enhance the quality of multimodal supervised fine-tuning data through the integration of vision priors, multi-expert assessments with domain-aware statistical fusion, GRPO-based critic training, and self-refinement mechanisms. Our extensive experiments demonstrate that VERITAS effectively denoises and refines training data, leading to significant improvements in downstream tasks, particularly in perception-centric benchmarks like OCR-VQA and MME. The lightweight GRPO critic achieves near-GPT-4o ranking fidelity while operating at a fraction of the computational cost, ensuring both efficiency and robustness. By systematically addressing issues such as factual errors and hallucinations in the data, VERITAS not only elevates the performance ceiling of subsequent large multimodal models but also maintains data diversity and richness. We believe that the release of the VERITAS pipeline, along with the refined dataset and model checkpoints, will facilitate future research in robust data curation and contribute to the development of more reliable and accurate multimodal language models.
\newpage
\section*{Limitations}
The limitations of our work are summarized as follows:

\noindent (1) The critique prompts and rewrite prompts that we have designed are relatively long. While this increases computational overhead, these comprehensive prompts provide richer contextual information, facilitating the model's more accurate understanding and generation of critique and rewritten content. These prompts can capture more nuanced semantic and syntactic information, thereby improving the quality of critiques and rewrites. Future work could consider how to optimize prompt design to maintain or enhance performance without significantly increasing computational costs.

\noindent (2) VERITAS leverages state-of-the-art LMMs (GPT-4o, Gemini-2.5-Pro, Doubao-1.5-pro) for critiques and refinements. Access to such powerful models may be restricted due to licensing, API limitations, or resource constraints. This dependency could pose challenges for broader adoption or replication of our results in different settings where these models are not readily accessible.

\bibliography{custom}

\appendix
\section{Additional Figures}

\begin{figure*}[h]
  \centering
  \includegraphics[width=0.95\linewidth]{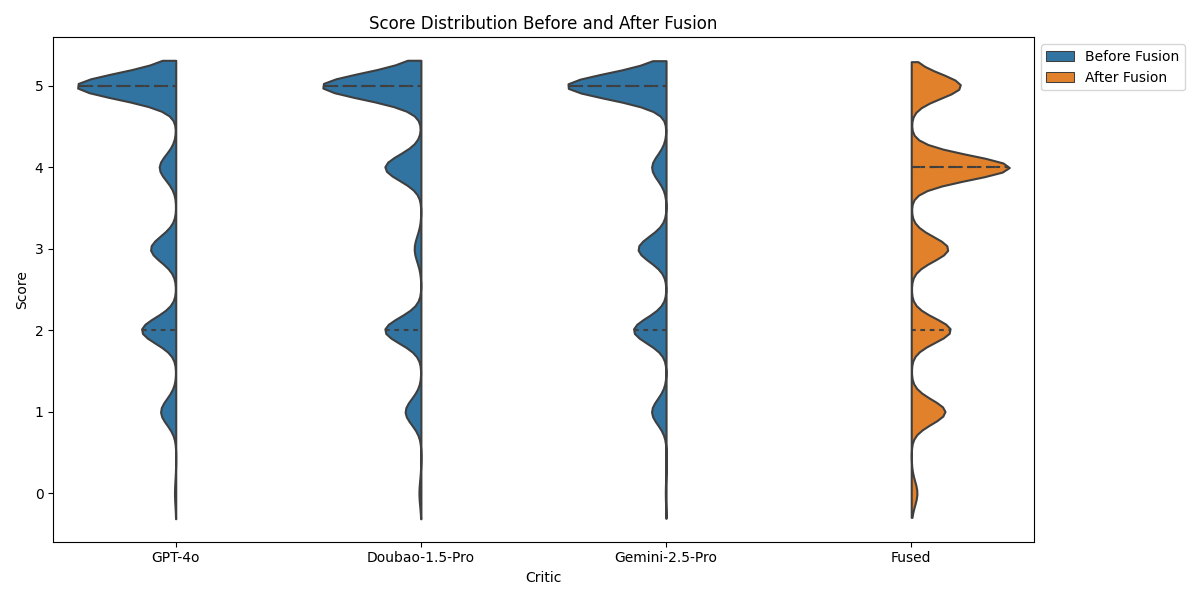}
  \caption{Score distributions from individual critics and after fusion. The fusion results in a more balanced distribution.}
  \label{fig:violin_scores}
\end{figure*}

\begin{figure*}[h]
  \centering
  \includegraphics[width=0.95\linewidth]{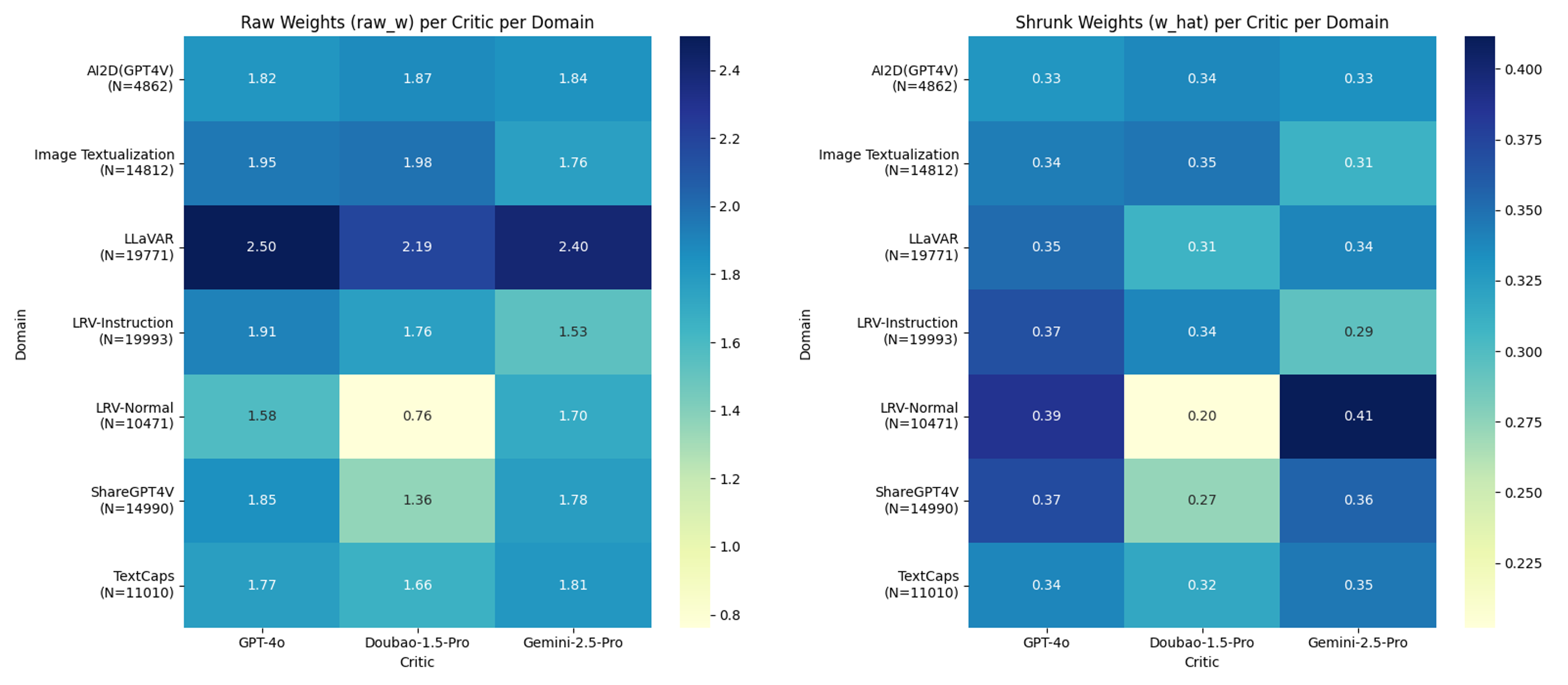}
  \caption{Critic weights before (raw) and after fusion across seven data sources. The fused weights adaptively adjust each critic's influence per domain.}
  \label{fig:critic_weights}
\end{figure*}

\section{Derivation and Analysis of Multi-Expert Fusion Method}
\label{sec:Derivation-Multi-Expert}
\subsection{Notation and Problem Setup}

In this appendix, we provide a detailed derivation and theoretical justification for the multi-expert fusion method presented in Algorithm~\ref{alg:fusion} of the main paper.

For each data source (domain) $d \in \mathcal{D}$, consider the following:

- Let $\mathbf{s}_d(n) = (s_{1,d}(n), s_{2,d}(n), s_{3,d}(n))^\top$ be the vector of raw scores assigned by the three expert critics for sample $n$ in domain $d$.
  
- Let $y_d(n)$ be the latent true score (e.g., a human-annotated score) for sample $n$ in domain $d$, assumed to have finite variance $\sigma_y^2$.

Our goal is to construct an estimator $\hat y_d(n)$ of the true score $y_d(n)$ by linearly combining the experts' scores:

\begin{equation}
\hat y_d(n) = \mathbf{w}_d^\top \mathbf{z}_d(n),
\end{equation}

where $\mathbf{w}_d = (w_{1,d}, w_{2,d}, w_{3,d})^\top$ are the weights for domain $d$, and \(\mathbf{z}_d(n)\) are the normalized scores, as defined below.

We aim to find weights \(\mathbf{w}_d\) that minimize the expected squared error (risk):

\begin{equation}
R = \mathbb{E}\left[ \left( \hat y_d(n) - y_d(n) \right)^2 \right].
\end{equation}

\subsection{Domain-Wise Z-Normalization}

To ensure comparability across different experts and domains, we perform z-score normalization of the raw scores within each domain:

\begin{equation}
z_{m,d}(n) = \frac{s_{m,d}(n) - \mu_{m,d}}{\sigma_{m,d}},
\end{equation}

where \(\mu_{m,d}\) and \(\sigma_{m,d}\) are the mean and standard deviation of expert \(m\)'s scores in domain \(d\), respectively.

By this normalization, the standardized scores \(z_{m,d}(n)\) satisfy:

\begin{equation}
\mathbb{E}[z_{m,d}] = 0, \quad \text{Var}[z_{m,d}] = 1.
\end{equation}

Any linear combination of these normalized scores will have expectation \(0\) if the weights sum to zero. However, since we aim to produce a meaningful aggregate score, we instead constrain the weights to sum to one:

\begin{equation}
\sum_{m=1}^3 w_{m,d} = 1.
\end{equation}

Although this introduces bias in the estimator, we correct for it later through the percentile rescaling step.

\subsection{Signal-to-Noise Ratio (SNR) Based Raw Weights}

Assuming that each expert's score is an unbiased estimator of the true score corrupted by noise, we model:

\begin{equation}
s_{m,d}(n) = y_d(n) + \eta_{m,d}(n),
\end{equation}

where \(\eta_{m,d}(n) \sim \mathcal{N}(0, \sigma_{m,d}^2)\) represents the noise in expert \(m\)'s score within domain \(d\).

The expected risk (mean squared error) of our estimator is then:

\begin{equation}
R = \mathbb{E}\left[ \left( \hat y_d(n) - y_d(n) \right)^2 \right] = \sum_{m=1}^3 w_{m,d}^2 \sigma_{m,d}^2,
\end{equation}

since the normalized scores \(z_{m,d}(n)\) are centered with unit variance.

To minimize \(R\), we set the weights proportional to the inverse of the variances:

\begin{equation}
w_{m,d} \propto \frac{1}{\sigma_{m,d}^2}.
\end{equation}

In practice, we estimate \(\sigma_{m,d}^2\) using the variance of the residuals (noise) within domain \(d\):

\begin{equation}
\sigma_{m,d}^2 \approx \text{Var}\left( s_{m,d}(n) - \bar s_d(n) \right),
\end{equation}

where \(\bar s_d(n)\) is the mean score for sample \(n\) across all experts in domain \(d\).

Thus, the raw weights are computed based on the signal-to-noise ratio (SNR):

\begin{equation}
\text{raw\_}w_{m,d} = \frac{\sigma_{m,d}}{\text{noise}_{m,d} + \epsilon},
\end{equation}

where \(\text{noise}_{m,d}\) is the standard deviation of the residuals \(r_{m,d}(n) = s_{m,d}(n) - \bar s_d(n)\), and \(\epsilon\) is a small constant to prevent division by zero.

\subsection{James--Stein Shrinkage Estimator}

The raw weights computed above may still have high variance, especially in domains with a small number of samples (\(N_d\) small). To address this, we apply James--Stein shrinkage, which shrinks the domain-specific weights toward the global mean weights, balancing bias and variance.

We compute the global mean weights:

\begin{equation}
\bar{\mathbf{w}} = \frac{1}{|\mathcal{D}|} \sum_{d \in \mathcal{D}} \mathbf{w}_d,
\end{equation}

and then apply shrinkage:

\begin{equation}
\hat{\mathbf{w}}_d = \alpha_d \mathbf{w}_d + (1 - \alpha_d) \bar{\mathbf{w}},
\end{equation}

where the shrinkage factor is:

\begin{equation}
\alpha_d = \frac{N_d}{N_d + \lambda},
\end{equation}

with \(\lambda\) being a hyperparameter set to \(\lambda = 100\) in our implementation.

This results in the adjusted weights \(\hat{\mathbf{w}}_d\), which are a convex combination of the domain-specific weights \(\mathbf{w}_d\) and the global mean weights \(\bar{\mathbf{w}}\). The choice of \(\alpha_d\) ensures that in domains with large \(N_d\), we trust the domain-specific weights more, while in domains with small \(N_d\), we rely more on the global weights.

\paragraph{Risk Reduction Proof}

Substituting the shrinkage weights into the risk \(R\), we find that the expected risk under the shrinkage estimator is less than or equal to that under the raw weights:

\begin{equation}
\begin{aligned}
\Delta R_d & = R(\hat{\mathbf{w}}_d) - R(\mathbf{w}_d)\\
& =  - \frac{(1 - \alpha_d)^2}{|\mathcal{D}|} \sum_{m=1}^3 (w_{m,d} - \bar w_m)^2 \le 0,
\end{aligned}
\end{equation}

since \((1 - \alpha_d)^2 \ge 0\) and the squared differences are non-negative. Equality holds only when \(\mathbf{w}_d = \bar{\mathbf{w}}\). Thus, the James--Stein shrinkage estimator does not increase the risk and typically reduces it.

\subsection{Percentile Re-Projection (Rescaling to Target Range)}

After fusing the normalized scores using the adjusted weights, we obtain the estimated scores:

\begin{equation}
\hat z(n) = \sum_{m=1}^3 \hat w_{m,d} \, z_{m,d}(n).
\end{equation}

However, the distribution of \(\hat z(n)\) may not be standard normal due to the weighting and shrinkage. To map the fused scores back to the original scoring range [0, 5], we apply a percentile-based linear rescaling.

We compute the lower and upper quantiles (e.g., 5\% and 95\%) of the fused scores \(\hat z(n)\) across all samples, denoted as \(q_{\text{low}}\) and \(q_{\text{high}}\), respectively.

The final estimated scores are then:

\begin{equation}
\hat S(n) = 5 \times \text{clip} \left( \frac{\hat z(n) - q_{\text{low}}}{q_{\text{high}} - q_{\text{low}}}, \, 0, 1 \right),
\end{equation}

where \(\text{clip}(x, 0, 1)\) constrains \(x\) to the interval [0, 1].

This rescaling ensures that the fused scores \(\hat S(n)\) lie within the desired range [0, 5], maintains the ordering (monotonicity), and reduces the impact of outliers by capping the extreme values.

\section{Automated Error Injection in CLEVR Dataset to Create Answer Quality Tiers}
\label{sec:Error-Injection}
To evaluate our evaluator's performance across varying answer qualities in an out-of-domain (OOD) setting, we modified the CLEVR dataset by introducing errors to create three tiers of answer quality:

\begin{itemize}
    \item \textbf{High-quality answers (Tier H):} Original correct answers were kept unchanged.
    \item \textbf{Medium-quality answers (Tier M):} Minor errors were introduced to make answers partially correct or slightly ambiguous. Examples include:
    \begin{itemize}
        \item Adjusting numerical answers by $\pm1$ or $\pm2$ (e.g., changing "4." to "5.").
        \item Replacing colors with similar ones (e.g., changing "Green." to "Blue." or "Cyan.").
        \item Switching size attributes (e.g., changing "Large." to "Small.").
        \item Changing definite answers like "Yes." to uncertain responses like "Maybe." or "Cannot tell.".
        \item Reversing material attributes (e.g., changing "Rubber." to "Metal.").
        \item Modifying shapes to similar ones (e.g., changing "Cube." to "Sphere.").
    \end{itemize}
    \item \textbf{Low-quality answers (Tier L):} Clear errors were introduced by replacing correct answers with incorrect values from different categories. Examples include:
    \begin{itemize}
        \item Swapping numerical answers with colors or shapes (e.g., answering "Red." instead of "3.").
        \item Changing "Yes." to "No." or providing contradictory statements.
        \item Providing unrelated attributes (e.g., answering "Metal." when the question asks for a color).
        \item Introducing nonexistent attributes (e.g., answering "Triangle." or "Plastic.", which are not present in CLEVR).
        \item Adding irrelevant explanations to incorrect answers.
    \end{itemize}
\end{itemize}

This automated error injection approach allowed us to generate a test set with diverse answer qualities, facilitating a comprehensive evaluation of our model's ability to handle varying levels of correctness in an OOD context while using LaTeX-compatible formatting for documentation.

\end{document}